# OPTIMIZING CNN-BIGRU PERFORMANCE: MISH ACTIVATION AND COMPARATIVE ANALYSIS WITH RELU

Asmaa BENCHAMA and Khalid ZEBBARA

IMISR Laboratory, Faculty of Science AM, Ibn zohr University, Agadir, Morocco

## ABSTRACT

*Deep learning is currently extensively employed across a range of research domains. The continuous advancements in deep learning techniques contribute to solving intricate challenges. Activation functions (AF) are fundamental components within neural networks, enabling them to capture complex patterns and relationships in the data. By introducing non-linearities, AF empowers neural networks to model and adapt to the diverse and nuanced nature of real-world data, enhancing their ability to make accurate predictions across various tasks. In the context of intrusion detection, the Mish, a recent AF, was implemented in the CNN-BiGRU model, using three datasets: ASNM-TUN, ASNM-CDX, and HOGZILLA. The comparison with Rectified Linear Unit (ReLU), a widely used AF, revealed that Mish outperforms ReLU, showcasing superior performance across the evaluated datasets. This study illuminates the effectiveness of AF in elevating the performance of intrusion detection systems.*

## 1. INTRODUCTION

Machine and Deep learning models have demonstrated significant benefits in the realm of intrusion detection[1], [2], offering enhanced capabilities in identifying and mitigating cybersecurity threats. The inherent capacity of deep learning to automatically learn hierarchical representations from data enables these models to discern intricate patterns and anomalies indicative of potential intrusions. The ability to capture complex relationships in vast and dynamic datasets makes deep learning particularly effective in detecting sophisticated and evolving cyber threats. Activation Functions *(AF)* Activation functions are central to this process by introducing non-linearities to the model's computations. They enable neural networks to understand and adapt to the nuanced nature of cyber threats, facilitating the learning of non-linear mappings between input features and potential intrusion outcomes. Properly chosen AF enhances the efficiency of the learning process, ensuring that the model can effectively generalize and make accurate predictions, thereby bolstering the overall effectiveness of deep learning models in intrusion detection systems. AF are a crucial component of deep learning models, enhancing their capacity to learn, generalize, and adapt to the intricacies present in diverse datasets, ultimately improving the overall efficiency and effectiveness of the models.

This paper offers substantial contributions across the following dimensions:

 -Investigating the influence of Mish implemented on the CNN-BiGRU model across diverse datasets and conducting a comparative analysis with the Rectified Linear Unit *(ReLU)* function. The study examines the effects of Mish [3]on the CNN-BiGRU model. This exploration spans a range of datasets, allowing for a comprehensive understanding of Mish's impact across different data characteristics. The comparative analysis with the widely used ReLU function offers valuable perspectives on the effectiveness of Mish. By delving into these aspects, the paper contributes valuable knowledge to the field, aiding researchers and practitioners in making informed decisions about AF choices in similar neural network architectures, thereby aiding them in responding to the question of which AF to use.

- The assessment of the intrusion detection system's efficacy is enriched by employing a variety of datasets, including ASNM-TUN[4], ASNM-CDX [4], and the recently introduced HOGZILLA dataset [5], [6]. This strategic choice of diverse databases serves to heighten the study's relevance and practical applicability in real-world settings. The evaluation ensures a comprehensive understanding of the model's performance, encompassing a spectrum of intrusion scenarios.
- Bridge the existing gap in the literature and offer insights into how AF can effectively tackle security challenges.

The subsequent sections of this article are structured as follows: Section 2 provides a comprehensive review of related works. Section 3 elucidates the methodology and materials employed in this study. In Section 4, experiments are performed on three core datasets and thoroughly analyze the obtained results. Ultimately, Section 5 concludes our work by summarizing key findings and drawing insightful conclusions.

## 2. RELATED WORKS

AF [7] plays a crucial and significant role in the field of artificial intelligence, particularly in the application of neural networks across various domains, including intrusion and attack detection. These mathematical operations are fundamental to the training process of neural networks, shaping their ability to understand and represent intricate patterns in diverse datasets. The introduction of non-linearities through AF is essential for the model to effectively learn and adapt to complex relationships within the input data. In computer vision, AF contributes to image recognition, enabling neural networks to identify objects, features, and patterns in visual data. In natural language processing, they aid in understanding and interpreting textual information, facilitating tasks such as sentiment analysis and language translation. Additionally, in healthcare, AF are integral to models that diagnose medical conditions based on patient data. In the field of intrusion and attack detection, AF enhance the capability of neural networks to recognize patterns indicative of cyber threats, contributing to the efficacy of security systems in safeguarding against potential breaches [8].

Due to the scarcity of papers addressing this topic, we begin by analyzing existing research that explores the influence of AF on the performance of deep learning models across different domains. Subsequently, we shift our attention to investigating this matter within the field of intrusion detection, aiming to fill the gap in literatures and provide insights into the effectiveness of AF in addressing security challenges.

Bircanoğlu et al. [9]conducted an analysis of the impacts of AF in Artificial Neural Networks *(ANN)* on both regression and classification performance. The experiments revealed that ReLU stands out as the most successful AF for general purposes. Furthermore, in addition to ReLU, the Square function demonstrated superior results, particularly in image datasets. N. Narisetty et al. [10] experimented with various AF such as linear, Leaky ReLU *(LReLU)*, ELU, Hyperbolic Tangent *(TanH)*, sigmoid, and softplus, which were selected for both the hidden and output layers. Adam optimizer and Mean Square Error loss functions were utilized to optimize the learning process. Assessing the classification accuracies of these AF using the CICIDS2017 dataset with the SVM-RBF classifier revealed that ELU exhibited superior performance with minimal computational overhead. Specifically, it achieved an accuracy of 97.33%, demonstrating only negligible differences compared to other AF. Szandała and Tomasz [11] assess various AF, including swish, ReLU, LReLU, Sigmoid, and provide a comparison of frequently utilized AF in deep neural networks, ReLU achieved an accuracy of 71.79%, while LReLU performed with an accuracy of 72.95%. P.Ramachandran et al.[12] Validate the efficacy of the searches through empirical evaluation using the AF that yielded the best results, Experiments reveal that Swish, identified as the optimal AF, exhibits superior performance in deeper models across various demanding datasets. D.Kim et al. [13]assessed EELU, and experimental findings indicate that EELU outperformed traditional AF, including ReLU, Exponential Linear Units *(ELU)*, ReLU and ELU-like variations, Scaled ELU, and Swish, by achieving enhanced generalization performance and classification accuracy. Dubey et al. [14] introduce various AF for neural networks in deep learning, encompassing different classes such as Logistic Sigmoid and TanH-based, ReLU-based, ELU-based, and Learning-based functions. Ratnawati et al. [15]employed eleven AF, including Binary Step Function, TanHRe, Swish, SoftPlus, Exponential Linear Squashing (ELiSH), LReLU, TanH, ELU, ReLU, Hard Hyperbolic Function *(HardTanH)*, and Sigmoid. Experimental results indicate that ELU and TanHRe demonstrate superior performance in terms of average and maximum accuracy on the Extreme Learning Machine *(ELM)*. In the realm of environmental contamination, Syed et al. [16]introduced an enhanced architecture known as the "Enhanced CNN model." This modification involved incorporating six additional convolutional layers into the conventional convolutional network. Furthermore, the utilization of the Mish in the initial five layers contributed to an enhancement in detection performance.

Studies have investigated the performance of various AF in neural networks across different tasks and datasets. In particular, ReLU is commonly reported as one of the most successful AF for general purposes, as demonstrated by Bircanoğlu et al.[9]. However, other studies have identified alternative AF, such as ELU, Swish, and Mish, that exhibit superior performance under certain conditions.

N. Narisetty et al. [10]found that ELU outperformed other AF in classification accuracy on the CICIDS2017 dataset. Additionally, P. Ramachandran et al. [12]identified Swish as the optimal AF for deeper models across various datasets. D. Kim et al. [13]also reported that EELU achieved enhanced generalization performance and classification accuracy compared to traditional AF like ReLU and ELU. In comparison to previous studies that highlighted the effectiveness of ReLU, ELU, and other AF, our findings underscore Mish's superior performance in enhancing model accuracy across diverse intrusion detection datasets. This suggests that Mish may offer a promising alternative to ReLU and other traditional AF in similar security-related applications. Our study reveals that Mish consistently outperforms the widely reported success of ReLU, as noted in the cited works, in terms of both macro F1-score and accuracy across all utilized datasets for intrusion detection in deep learning.

Table 1. Summary of related works

| Ref | AF | Datasets | Achievement |
|---|---|---|---|
| [9] | ReLU, Linear, H.Sigm, Softsign, ELU, SeLU, Swish, Softplus, sigmoid, TanH, kare on ANN | Boston, Ames, Reuters, Fashion MNIST, MNIST, CIFAR10 and IMDB Image datasets | The comparative analysis indicates that ReLU performs exceptionally well across all image datasets. |
| [10] | linear, LReLU, ELU, TanH, sigmoid, and softplus with the SVM-RBF classifier | CICIDS2017 dataset | ELU demonstrated outstanding performance while requiring minimal computational resources. |
| [11] | Swish, ReLU, LReLU, Sigmoid on CNN | CIFAR-10 image dataset | ReLU and LReLU emerged as the most successful, as all other networks achieved task completion with less than 70% accuracy |
| [12] | Swish, ReLU and Softplus | CIFAR10, CIFAR100 and ImageNet classification | Swish yields superior performance compared to Softplus and ReLU. |
| [13] | EELU, ReLU, ELU, EPReLU and Swish on A simple CNN model and VGG16 | CIFAR10, CIFAR100, ImageNet, and Tiny ImageNet classification | EELU demonstrates superior performance compared to ReLU, ELU, EPReLU, and Swish. |
| [14] | Surveying the performance analysis of AF on Deep Neural Networks. | Image, Text and Speech data | TanH and SELU AF are identified as more effective for language translation, alongside PReLU, LiSHT, SRS, and PAU. For speech recognition tasks, it is recommended to utilize PReLU, GELU, Swish, Mish, and PAU AF. |
| [15] | Sigmoid, Swish, ELiSH, HardTanH, ReLU, TanHRe, ELU, SoftPlus, and LReLU on Extreme Learning Machine. | Compounds that are active based on their SMILES structure | The mean accuracy can achieve 80.56% with ELUs AF, while the highest accuracy of 88.73% is attained with TanHRe. |
| [16] | Mish in the initial five layers into the conventional convolutional network | Detection of contamination and the use of a time-of-flight sensor to ensure safe interaction between machines and the environment. | Mish contributes to enhancing the detection performance of the model. |
| - | Mish and ReLU over hybrid deep learning model CNN-BiGRU (our contribution) | Hogzilla dataset, ASNM-TUN and ASNM-CDX2006 In the field of intrusion detection. | Mish demonstrates superior performance compared to ReLU across all datasets utilized, effectively enhancing overall performance. We must note that we observed a scarcity of papers thoroughly reviewing the AF utilized by neural networks, especially in the field of intrusion detection |

It is noteworthy to highlight that there is a limited body of research especially in the field of attack detection and cyber-security that explicitly delves into the effects and implications of AF in machine learning and neural network applications. Despite the pivotal role AF plays in shaping the learning capabilities of models, a comprehensive exploration of their influences on performance and convergence is relatively sparse in existing literature particularly in the security and attack detection domain. This underlines the need for more in-depth investigations into the nuances and significance of AF in different contexts to enhance our understanding and optimization of neural network architectures.

## 3. METHODOLOGY AND MATERIALS

In this section, we embark on a thorough exploration of the AF commonly employed in the realm of deep learning. We delve into various AF, elucidating their characteristics, advantages, and limitations to provide readers with a comprehensive understanding of their role in model development.

Furthermore, we introduce our CNN-BiGRU model, a hybrid architecture that combines Convolutional Neural Networks *(CNN)* with Bidirectional Gated Recurrent Units *(BiGRU)*. This architecture is tailored to leverage the strengths of both CNN and BiGRU, offering enhanced performance in tasks such as sequence modeling and feature extraction.

Additionally, we elucidate the Synthetic Minority Over-sampling Technique *(SMOTE)*[17] method utilized in our study to address imbalanced data distribution. Imbalanced datasets are ubiquitous in real-world applications, posing challenges for Deep learning models. By employing SMOTE, we aim to mitigate the adverse effects of class imbalance by generating synthetic samples for minority classes, thus fostering a more balanced training set and improving model performance.

Through this comprehensive overview and detailed description of our model architecture and data preprocessing techniques, we equip readers with the necessary knowledge and insights to navigate the complexities of AF selection, model design, and data preprocessing in deep learning applications.

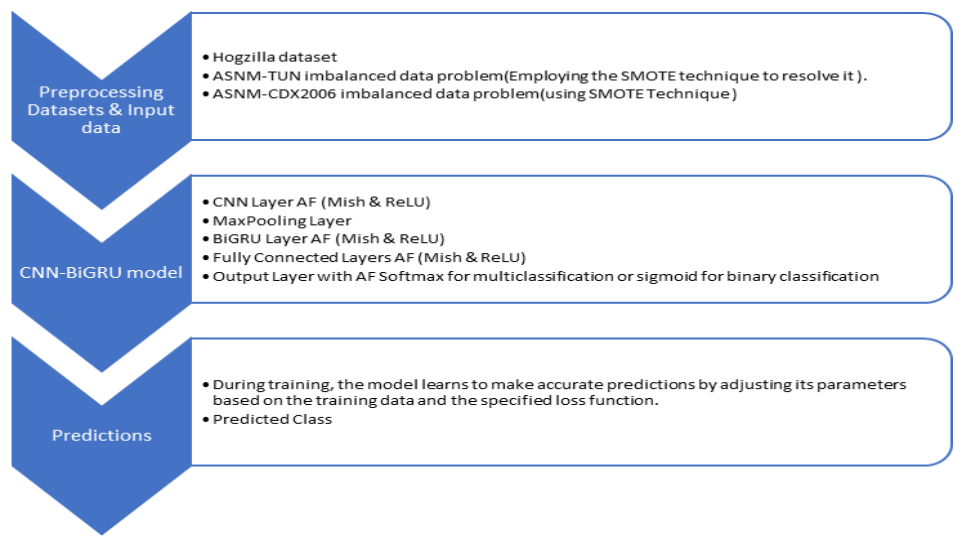

Figure 1. Architecture Overview of CNN-BiGRU Model

The table 2 outlines the components of the CNN-BiGRU model, including the layers, AF, and parameters such as the number of filters, kernel size, pool size, number of units, and number of classes. Additionally, it specifies the simulation environment.

Table 2. Architecture Overview and Simulation Environment

| Component | Description |
|---|---|
| Input Data | Hogzilla dataset, ASNM-TUN, ASNM-CDX2006 |
| CNN Layer | AF: ReLU, Mish |
|  | Number of Filters: 32 |
|  | Kernel Size: 3x3 |
| MaxPooling Layer | Pool Size: 2x2 |
| BiGRU Layer | AF: ReLU, Mish |
|  | Number of Units: 64 |
| Fully Connected Layers | AF: ReLU, Mish |
|  | Number of Units: 128 |
| Output Layer | AF: Softmax for multiclass classification, Sigmoid for binary classification |
| Simulation Environment | Python Version: 3.8.5 |
|  | TensorFlow Version: 2.5.0 |
|  | Keras Version: 2.5.0 |

### 3.1. Activation Functions

AF serves as mathematical operations applied to the output of each neuron in a neural network, introducing non-linearities that enable the model to learn and represent intricate patterns in data. They hold a central role in the training process, influencing the network's capacity to capture diverse relationships within input data. An in-depth overview of commonly utilized AF in deep learning is outlined as follows:

#### 3.1.1. Sigmoid Function

$$Sigmoid(x) = \frac{1}{1+e^{-x}} \quad (1)$$

- $x$ is the input to the function.
- $e$ is the mathematical constant approximately equal to 2.71828.

Commonly employed in the output layer for binary classification problems. Less common in hidden layers due to issues like vanishing gradients. The sigmoid function outputs values in the range of 0 to 1. Frequently employed in the output layer of binary classification models, its purpose is to generate a probability indicating the likelihood that an input pertains to a specific class. The sigmoid function transforms any real number into a range [0, 1], making it useful for binary decision problems.

---

[1] sigmoid formula available at https://www.analyticsvidhya.com/

International Journal of Computer Networks & Communications (IJCNC) Vol.16, No.3, May 2024

### 3.1.2. Hyperbolic Tangent (TanH)

TanH function is an AF commonly used in machine learning and neural networks. Its formula is defined as:

$$Tanh(x) = \frac{e2x - 1}{e2x + 1} \quad (2)$$

Effective in hidden layers for capturing more complex patterns. The TanH function outputs values in the range of -1 to 1. Similar to the sigmoid function, the TanH function is particularly effective in the hidden layers for capturing more intricate patterns in the data. The TanH function introduces non-linearity and is especially useful when the input data has negative values.

### 3.1.3. Rectified Linear Unit (Relu)

$$f(x) = max(0, x); \quad \text{Range: } [0, \infty). \quad (3)$$

Widely used, replaces negative values with zero, introducing non-linearity. This function might encounter the issue known as the "dying ReLU" problem, leading to neurons becoming inactive.

### 3.1.4. Leaky Relu

$$f(x) = max(\alpha x, x) \quad (4)$$

where α is a small positive constant; Range: (-∞, ∞).
Mitigates the "dying ReLU" problem by permitting a small gradient for negative values.

### 3.1.5. Parametric Relu (Prelu)

Extension of LReLU where during training, the slope of the negative segment is learned.

### 3.1.6. Exponential Linear Unit (Elu)

$$f(x) = x \text{ if } x > 0 \quad \text{Else} \quad \alpha \cdot (ex - 1) \quad (5)$$

where α is a small positive constant.

Similar to ReLU with smoother transitions for negative values, aiming to mitigate the dying ReLU problem.

### 3.1.7. Mish Function

$$f(x) = x \cdot tanh(softplus(x)) \quad (6)$$

---

[2] [3] [4] Hyperbolic tangent, ReLU, Leacky ReLU formula available at https://www.analyticsvidhya.com/blog/

[5] PReLU, ELU formula available at https://www.analyticsvidhya.com/blog/

Mish is a newer AF proposed to address some of the limitations of ReLU.

It has a smooth curve and is non-monotonic, providing a more continuous gradient throughout the range of inputs. Mish has shown promising results in certain scenarios, potentially helping mitigate the dying ReLU problem and providing improved generalization.

### 3.1.8. Softmax Function

$$Softmax(z)i = \frac{e^{z_i}}{\sum_{j=1}^{K} e^{z_j}} \quad (7)$$

The softmax function is commonly used in the output layer of a neural network for multi-class classification problems. $z$ is a vector of raw scores (logits) for each class.
K is the total number of classes.

Softmax($z$)i is the computed probability for class i.

The softmax function calculates the exponential of each score and normalizes it by dividing it by the sum of the exponentiated scores across all classes. This normalization guarantees that the probabilities obtained add up to 1, which is advantageous for multi-class classification tasks. The class with the highest probability is commonly selected as the predicted class.

These AF serve distinct purposes and are chosen based on specific task requirements, network architecture, and data characteristics. Choosing the right activation function is a crucial element in crafting a successful neural network.

We Implemented the Mish throughout the CNN-BiGRU, the study involves a comparison between Mish and ReLU. The assessment includes an evaluation of accuracy, precision, recall, and other relevant metrics on datasets for intrusion detection. This comprehensive empirical analysis aims to provide insights into the efficacy of Mish and ReLU in capturing spatial and temporal patterns for enhancing the intrusion detection capabilities of the CNN-BiGRU model or similar hybrid deep learning model.

### 3.2. CNN-BIGRU Model

The CNN-BiGRU model operates through a two-step process to effectively detect network intrusions. Firstly, a One-Dimensional CNN is employed to comprehensively extract local features from the traffic data, ensuring the capture of pertinent and specific information. This initial step is crucial for identifying relevant patterns within the dataset. Following this, BiGRU is employed to capture time series features, offering a comprehensive grasp of temporal patterns within the traffic data. The bidirectional aspect of BiGRU enables the model to incorporate information from both preceding and subsequent time steps, thereby augmenting its temporal comprehension.

---

[6]Mish formula available at
https://www.scirp.org/journal/paperinformation?paperid=114024#:~:text=The%20formula%20of%20the%20Mish,is%20proposed%20by%20Yoshua%20Bengio.
[7]Solftmax function available at
https://www.engati.com/glossary/softmaxfunction#:~:text=The%20softmax%20formula%20computes%20the,output%20of%20the%20softmax%20function.

The combination of CNN[18] and BiGRU[19] in the model ensures the extraction of a rich set of information hidden within the traffic data. This comprehensive approach significantly enhances the model's ability to discern intricate patterns and anomalies associated with network intrusion, making it well-suited for complex detection tasks. To further optimize the model, a Global Average Pooling Layer is employed in the final stage instead of a fully connected layer. This strategic replacement serves to reduce network parameters and computational complexity, mitigating the risk of overfitting. By incorporating these elements, the CNN-BiGRU model emerges as a robust solution for intrusion detection, capable of effectively handling diverse patterns and variations within network traffic data.

### 3.3. Overview of the SMOTE Technique

SMOTE aims to rectify class imbalance by generating synthetic samples for the minority class, thereby furnishing the model with a more balanced and representative dataset for training. The SMOTE algorithm operates by generating synthetic instances of the minority class along the line segments connecting existing minority class instances. It picks a sample from the minority class, locates its k nearest neighbors, and creates synthetic samples along the lines connecting the sample and its neighbors. Sampling Strategy: Involves determining the ratio of the number of synthetic samples to be generated to the number of existing minority class samples. The application of SMOTE is crucial, particularly when dealing with datasets presenting a considerable number of minority classes. In our case, we applied the SMOTE technique on two datasets, ASNM-TUN and ASNM-CDX, for featuring numerous minority classes.

In summary, SMOTE serves as a valuable tool for enhancing the performance of machine learning and deep learning models dealing with imbalanced datasets, especially when working with datasets featuring numerous minority classes. By fostering a more equitable distribution of classes, SMOTE contributes to the model's improved generalization and accuracy, especially for minority class instances.

The formula for SMOTE is provided as follows:

$$xsynth = xi + \lambda \cdot (xni - xi) \text{ and } 0 \leq \lambda \leq 10 \quad (8)$$

In the formulas, $xi$ represents a minority class instance, $xni$ represents one of its k-nearest neighbors, and λ is a random value between 0 and 1. The process is repeated for multiple instances in the minority class.

## 4. EXPERIMENTS AND RESULTS

In our experimental setup, we conducted a comprehensive evaluation of the Mish and ReLU within the CNN-BiGRU model using three distinct datasets: ASNM-TUN, ASNM-CDX, and HOGZILLA, all tailored for intrusion detection scenarios. The Mish, a recent addition to the repertoire of AF, was implemented alongside the widely used ReLU function. Our assessment focused on key performance metrics such as accuracy, precision, and recall, providing a detailed comparison of the two AF in capturing spatial and temporal patterns indicative of network intrusions. The experiments were conducted on a one-dimensional CNN and BiGRU. The results, presented shed light on the comparative effectiveness of Mish and ReLU in enhancing the intrusion detection capabilities of the CNN-BiGRU model across different datasets.

---

(8) Smote formula available at https://www.researchgate.net/figure/Equations-for-1-synthetic-minority-oversampling-technique-2-inverse-document_fig2_331002009

## 4.1. Datasets

### 4.1.1. Hogzilla Dataset

The Hogzilla Dataset[1] is a composite of network flows extracted from the CTU-13 Botnet and the ISCX 2012 IDS datasets. Every flow in the dataset contains 192 behavioral features. This compilation results in a dataset that encompasses the behavioral information of Botnets identified in the CTU-13 dataset, as well as normal traffic from the ISCX 2012 IDS dataset.

Pre-processing of the original HOGZILLA dataset involved transforming attack labels into their corresponding attack classes: 'ACCEPTABLE', 'Unrated' and 'UNSAFE'. The classification of attack labels into their respective classes is as follows:

['Acceptable', 'Safe'] → Acceptable;
 ['Unrated', 'Fun'] → Unrated;
['Unsafe', 'Dangerous'] → Unsafe.

The primary steps undertaken during pre-processing include:

-Selection of numeric attribute columns from the dataset.
-Application of a standard scaler to normalize the selected numeric attributes.
-Label encoding (0, 1, 2) for multi-class labels ('Acceptable', 'Unrated', 'Unsafe').

-Creation of a dataframe containing only the numeric attributes of the multi-class dataset and the encoded label attribute.

-Identification of attributes with a correlation greater than 0.5 with the encoded attack label attribute using Pearson correlation coefficient.

-Selection of attributes based on the identified correlation, contributing to the final dataset configuration.

Each class within the dataset consists of the following number of flows: Unrated: 5657 ; Unsafe: 4546 ;Acceptable: 2629.

### 4.1.2. Asnm-Tun Dataset

Advanced Security Network Metrics & Tunneling Obfuscations [2]dataset incorporates instances of malicious traffic with applied tunneling obfuscation techniques, and its creation dates back to 2014.

The ASNM-TUN [2]dataset comprises four distinct label types, organized in ascending order of granularity as outlined in the following enumeration:

-The binary label, referred to as label_2, indicates whether a given record represents a network attack or not.
-The three-class label, labeled as label_3, discerns between legitimate traffic (symbol 3) and direct and obfuscated network attacks (symbols 1 and 2).

---

[1] Hogzilla dataset available at https://ids-hogzilla.org/dataset/
[2] ASNM-TUN dataset available at http://www.fit.vutbr.cz/~ihomoliak/asnm/ASNM-TUN.html

- The label labeled as label_poly is structured in two parts: three-class label, and acronym of network service.
- The last label, labeled as label_poly_s, is structured of 3 parts: three-class label, acronym of network service, and employed network modification technique.
- The label, shares a similar interpretation with the previous one but additionally introduces the employed network modification technique, identified by a letter from the listing.

Each category in the The dataset consists of the following number of flows: ATTACK 177 instances;SAFE; 130 instances;OFFUSCATED 87 instances.

We employed the SMOTE technique to resolve the class imbalance problem in the ASNM-TUN dataset. This imbalance arose due to the varying quantities of flows in each category, with 177 instances in ATTACK, 130 instances in SAFE, and 87 instances in OFFUSCATED.

### 4.1.3. Asnm-Cdx2009 Dataset

Advanced The utilized dataset originates from the ASNM datasets[20], specifically named ASNM-CDX-2009[3] , and was curated by the National Security Agency of the United States of America (NSA).The designation "CDX" stands for "Cyber Defense Exercise[21]." Described by the NSA, the Cyber Defense Exercise (CDX) is an annual simulated real-world educational event designed to challenge university students to build secure networks and defend against adversarial attacks. ASNM, standing for Advanced Security Network Metrics, represents a compilation of network data characterizing TCP connections with diverse attributes. These datasets were crafted to cater to the requirements of traffic analysis, threat detection, and recognition. chosen dataset specifically focuses on traffic conducted through the TCP protocol. Although the original CDX-2009 collection comprises around four million entries, the ASNM originators[22]have restricted it to around 5713 categorized connections.

The entries within the ASNM-CDX-2009 file are classified and do not encompass data transmitted in IP packets. Each entry is characterized by two labels: "label_2," which denotes whether the entry pertains to a buffer overflow attack, and "label_poly," featuring a binary-descriptive structure. The first segment of "label_poly" indicates whether the traffic is deemed safe or unsafe, denoted by the values zero and one, respectively. The second segment specifies the service associated with the traffic, with three defined service descriptions: apache, postfix, and others. For instance, an entry labeled 0_postfix signifies a secure connection related to email service.

Data from the ASNM-CDX-2009 database is accessible for download from the internet. Furthermore, the network traffic records forming the basis of the CDX-2009 dataset are publicly available.

Each class within the used dataset consists of the following number of flows:
SAFE 5692 instances; UNSAFE  43instances.

The dataset we utilized contained a substantial class imbalance, with the majority class, labeled as SAFE, comprising 5692 instances, and the minority class, labeled as UNSAFE, having only 43 instances. SMOTE was applied as a data augmentation.

---

[3] ASNM-CDX2009 dataset available at http://www.fit.vutbr.cz/~ihomoliak/asnm/ASNM-CDX-2009.html

International Journal of Computer Networks & Communications (IJCNC) Vol.16, No.3, May 2024

## 4.2. Evaluation Metrics

The assessment metrics for the intrusion detection model in network security encompass precision, accuracy, recall, and F1-score. These metrics are calculated based on the information provided by the confusion matrix, which classifies the model's classification outcomes into four categories: True Positive (TP), True Negative (TN), False Positive (FP), and False Negative (FN). Here are the definitions for each term:

-True Positive (TP): Instances correctly identified as positive.
-True Negative (TN): Instances correctly identified as negative.
-False Positive (FP): Instances incorrectly identified as positive.
-False Negative (FN): Instances incorrectly identified as negative.

These metrics play a crucial role in assessing the model's performance and effectiveness in identifying network security intrusions.

## 4.3. Results

Our experimental evaluations involve three datasets, specifically ASNM-TUN, ASNM-CDX2009, and the recent HOGZILLA. These datasets encompass a diverse range of attacks and properties, providing a comprehensive evaluation ground. In tackling the issue of imbalanced data in ASNM-TUN and ASNM-CDX2009, we employed the SMOTE technique, which proved effective in generating synthetic samples and resolving the imbalance issue. The strategic selection of these three datasets ensures an efficient evaluation across a large and varied dataset. This exploration aims to provide intricate insights into how AF contributes to the overall performance of the model, shedding light on their varying impacts and effectiveness in optimizing accuracy.

### 4.3.1. Hogzilla Dataset Results

Using the Mish, the CNN-BiGRU model demonstrates as shown in Table 3 an accuracy of 98.72% and achieves a macro F1- score of 98.44% on the Hogzilla dataset.

Table 3. Mish results on Hogzilla Dataset

| Num | Class | Precision | Recall | F1-Score |
|---|---|---|---|---|
| 0 | Acceptable | 98.60% | 95.63% | 97.09% |
| 1 | Unrated | 99.01% | 99.57% | 99.29% |
| 2 | Unsafe | 98.43% | 99.47% | 98.95% |
| Metric | Accuracy | Macro Precision | Macro Recall | Macro F1-Score |
| Value | 98.72% | 98.68% | 98.22% | 98.44% |

The utilization of the ReLU in the CNN-BiGRU model on the Hogzilla dataset as shown in Table 4 yields an accuracy of 98.32% and a macro F1-score of 98.06%.

80

Table 4. ReLU results on Hogzilla Dataset

| Num | Class | Precision | Recall | F1-Score |
|---|---|---|---|---|
| 0 | Acceptable | 98.44% | 95.17% | 96.78% |
| 1 | Unrated | 98.93% | 98.72% | 98.83% |
| 2 | Unsafe | 97.51% | 99.65% | 98.57% |
| Metric | Accuracy | Macro Precision | Macro Recall | Macro F1-Score |
| Value | 98.32% | 98.29% | 97.85% | 98.06% |

Mish achieved a slightly higher accuracy (98.72%) compared to the ReLU (98.32%). This suggests that Mish is more effective in capturing complex patterns in the Hogzilla dataset, leading to better overall classification performance. The Mish resulted in a higher macro F1-score (98.44%) compared to the ReLU (98.06%).

### 4.3.2. ASNM-TUN Dataset Results

Using the Mish activation When employing the Mish on the CNN-BiGRU model with the ASNM-TUN dataset, as shown in Table 5 the outcomes include an accuracy of 92.31% and a macro F1-score of 91.57%.

Table 5. Mish results on ASNM-TUN Dataset

| Num | Class | Precision | Recall | F1-Score |
|---|---|---|---|---|
| 0 | Safe | 78.95% | 93.75% | 85.71% |
| 1 | Obfuscated | 99.99% | 88.24% | 93.75% |
| 2 | Attack | 96.77% | 93.75% | 95.24% |
| Metric | Accuracy | Macro Precision | Macro Recall | Macro F1-Score |
| Value | 92.31% | 91.91% | 91.91% | 91.57% |

Utilizing the ReLU in the CNN-BiGRU model on the ASNM-TUN dataset as shown in Table 6 yields results comprising an accuracy of 87.79% and a macro F1-score of 86.48%.

Table 6. ReLU results on ASNM-TUN Dataset

| Num | Class | Precision | Recall | F1-Score |
|---|---|---|---|---|
| 0 | Safe | 85.71% | 97.67% | 91.30% |
| 1 | Obfuscated | 80.77% | 77.78% | 79.25% |
| 2 | Attack | 92.86% | 85.25% | 88.89% |
| Metric | Accuracy | Macro Precision | Macro Recall | Macro F1-Score |
| Value | 87.79% | 86.45% | 86.90% | 86.48% |

The model with the Mish achieved a higher accuracy (92.31%) compared to the ReLU (87.79%). Similar to accuracy, the Mish resulted in a higher macro F1-score (91.57%) compared to the ReLU (86.48%)

**4.3.3. ASNM-CDX2009 Dataset Results**

Using When employing the Mish in the CNN-BiGRU model with the ASNM-CDX2009 dataset, the obtained results as shown in Table 7 include an accuracy of 99.79% and a macro F1-score of 92.80%.

Table 7. Mish results on ASNM-CDX2009 Dataset

| Num | Class | Precision | Recall | F1-Score |
|---|---|---|---|---|
| 0 | Safe | 99.79% | 99.99% | 99.89% |
| 1 | Unsafe | 99.99% | 75.00% | 85.71% |
| Metric | Accuracy | Macro Precision | Macro Recall | Macro F1-Score |
| Value | 99.79% | 99.89% | 87.50% | 92.80% |

Applying the ReLU in the CNN-BiGRU model to the ASNM-CDX2009 dataset as shown in Table 8 produces results, including an accuracy of 99.58% and a macro F1-score of 84.51%.

Table 8. ReLU results on ASNM-CDX2009 Dataset

| Num | Class | Precision | Recall | F1-Score |
|---|---|---|---|---|
| 0 | Safe | 99.68% | 99.89% | 99.79% |
| 1 | Unsafe | 81.82% | 60.00% | 69.23% |
| Metric | Accuracy | Macro Precision | Macro Recall | Macro F1-Score |
| Value | 99.58% | 90.75% | 79.95% | 84.51% |

The model with the Mish achieved an extremely high accuracy (99.79%), slightly higher than the ReLU (99.58%). The Mish resulted in a higher macro F1- score (92.80%) compared to the ReLU (84.51%).

**4.4. Discussion**

The results of our study shed light on the performance of Mish and ReLU in deep learning models for intrusion detection. Before delving into the numerical results, we succinctly analyze the characteristics of Mish and ReLU:

Table 9. Comparison of Mish and ReLU

| Aspect | Mish | ReLU |
|---|---|---|
| AF Type | Non-linear | Non-linear |
| Characteristics | Smooth, bounded | Piecewise linear |
| Advantages | Improved convergence | Simplicity, faster |
| Disadvantages | Complexity, computation | Vanishing gradient |
| Common Usage | Recent research trend | Widely used in practice |

Mish, being a recent research trend, exhibits characteristics such as smoothness and boundedness, contributing to improved convergence during training. However, Mish may entail higher computational complexity compared to ReLU. On the other hand, ReLU, widely used in practice, offers simplicity and faster computation but is susceptible to issues such as vanishing gradients. This comparison highlights the trade-offs and considerations associated with choosing between Mish and ReLU for intrusion detection tasks. Our numerical results further complement this analysis, providing quantitative evidence of their performance on various datasets.

Mish demonstrates consistent superiority across all three datasets in terms of accuracy and macro F1-score. Mish effectively captures intricate patterns and relationships in all datasets. On the other hand, ReLU Falls short compared to Mish in capturing nuanced patterns, as evidenced by lower macro F1-score in all datasets.

Table 10. Summary of Mish and ReLU results

| Dataset | Metric | Mish (%) | ReLU (%) | Difference (%) |
|---|---|---|---|---|
| ASNM-TUN | Precision | 91.91 | 86.45 | +5.46 |
| | Recall | 91.91 | 86.90 | +5.01 |
| | F1-Score | 91.57 | 86.48 | +5.09 |
| | Accuracy | 92.31 | 87.79 | +4.52 |
| ASNM-CDX2006 | Precision | 99.89 | 90.75 | +9.14 |
| | Recall | 87.50 | 79.95 | +7.55 |
| | F1-Score | 92.80 | 84.51 | +8.29 |
| | Accuracy | 99.79 | 99.58 | +0.21 |
| Hogzilla | Precision | 98.68 | 98.29 | +0.39 |
| | Recall | 98.22 | 97.85 | +0.37 |
| | F1-Score | 98.44 | 98.06 | +0.38 |
| | Accuracy | 98.72 | 98.32 | +0.40 |

Across the ASNM-TUN, ASNM-CDX2006, and Hogzilla datasets, Mish as shown in Table 10 tends to achieve higher precision compared to ReLU, with differences of +5.46%, +9.14%, and +0.39%, respectively, This suggests that Mish is better at correctly identifying positive instances with very few false positives.. Mish shows a slight advantage in recall values, with differences of +5.01%, +7.55%, and +0.37%, respectively, indicating that Mish is better at capturing true positive instances. In terms of F1-score, Mish consistently outperforms ReLU, with differences of +5.09%, +8.29%, and +0.38%, respectively, indicating a better balance between precision and recall for Mish. Furthermore, Mish achieves slightly higher accuracy compared to ReLU across all datasets, with differences of +4.52%, +0.21%, and +0.40%, respectively. These findings highlight Mish's overall superior performance across various evaluation metrics and datasets, suggesting its effectiveness as an AF for intrusion detection tasks.

Mish, being a smooth and bounded AF, may help prevent exploding gradients and facilitate smoother optimization compared to ReLU, which is piecewise linear and prone to issues such as vanishing gradients. This could contribute to Mish's superior performance in capturing intricate patterns in the data and making more accurate predictions. Mish's smoothness and boundedness could be advantageous in scenarios where the data exhibits complex and non-linear relationships, such as intrusion detection tasks. Mish's ability to maintain non-zero gradients even for negative inputs may help mitigate the problem of dying neurons, which is a common issue with ReLU.

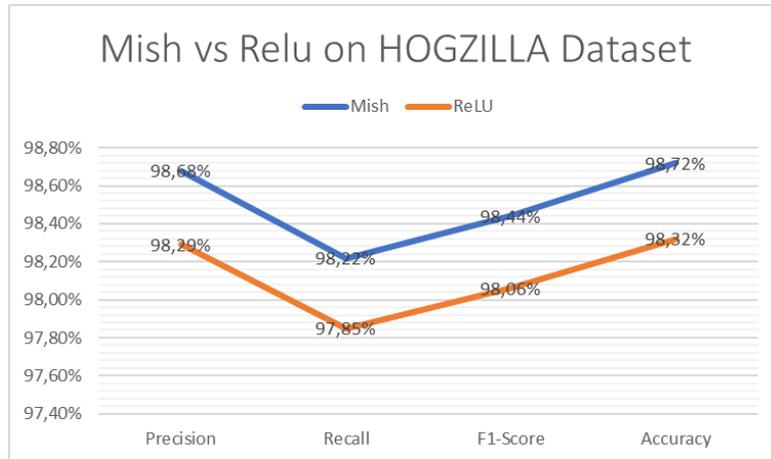

Figure 2. Mish and ReLU performance on Hogzilla

Figure 2 illustrates the performance comparison between Mish and ReLU on the Hogzilla dataset across four key metrics macro average values: precision, recall, F1-score, and accuracy. Each metric provides insights into the effectiveness of Mish and ReLU in accurately classifying. The graph visually depicts Mish's superior performance over ReLU across multiple metrics based on Hogzilla dataset.

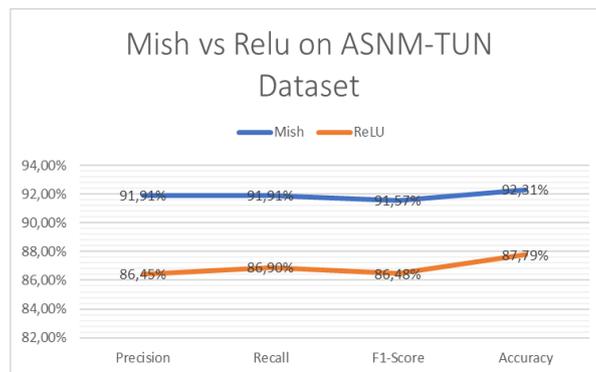

Figure 3. Mish and ReLU performance on ASNM-TUN

Figure 3 visually highlights the notably higher difference between Mish and ReLU across all metrics, emphasizing Mish's superior performance over ReLU based on the ASNM-TUN dataset.

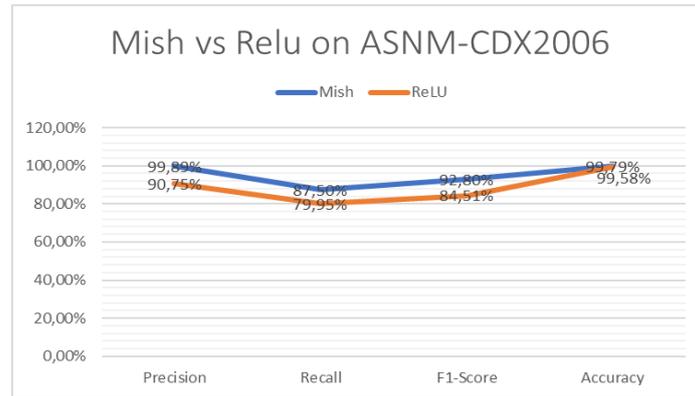

Figure 4.  Mish and ReLU performance on ASNM-CDX2006

In figure 4 Mish demonstrates superiority in terms of F1-score, accuracy, recall, and precision, the difference, while significant, is not as pronounced as observed in the Hogzilla and ASNM-TUN datasets in terms of the accuracy metric.

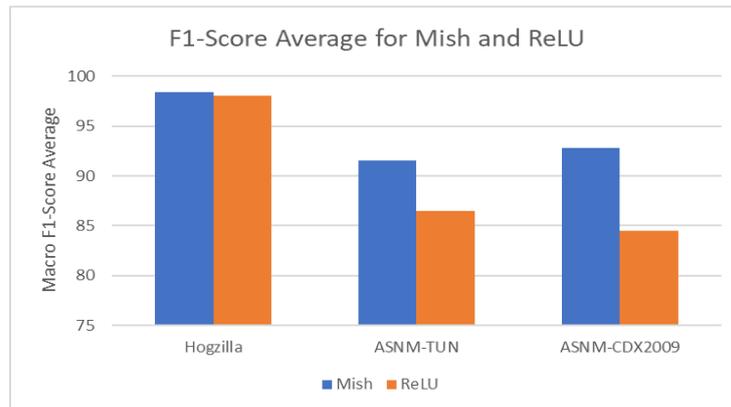

Figure 5.  Macro F1-Score average of Mish and ReLU on all datasets

Figure 5 visually demonstrates the superiority of Mish over ReLU based on the macro F1-score, a metric crucial for evaluating the overall effectiveness of classification models. The macro F1-score considers both precision and recall across all classes, providing a balanced assessment of the model's performance.

The results indicate that Mish generally outperforms ReLU across various metrics and datasets as shown in Figure 2,3,4 and 5, suggesting its potential as a more effective AF for intrusion detection tasks. Mish's smoother and bounded nature appears to contribute to its superior performance in capturing complex patterns and making accurate predictions compared to ReLU.

## 5. CONCLUSION

In conclusion, this study underscores the important role of AF in improving the performance of intrusion detection systems within the context of deep learning, specifically focusing on the CNN-BiGRU model. Leveraging three diverse datasets (ASNM-TUN, ASNM-CDX, and HOGZILLA). The findings consistently demonstrate Mish's superior performance across all evaluated datasets in terms of both accuracy and macro F1-score. Mish is a non-linear activation function created to mitigate certain drawbacks of conventional activation functions such as ReLU. It introduces a smooth, differentiable function that can help mitigate issues like dead neurons and vanishing gradients. The better performance of Mish in this scenario suggests that the model benefits from the characteristics introduced by Mish, in capturing more features in the data. These findings offer noteworthy conclusions about how AF enhances the adaptability and pattern-recognition abilities of neural networks.

In our future research efforts, our aim is to expand upon the investigation into the impact of AF beyond the CNN-BiGRU architecture. We aim to explore their influence on a broader range of neural network architectures, considering variations in model structures and complexities. Additionally, we plan to conduct experiments using diverse datasets beyond the ones examined in this study (ASNM-TUN, ASNM-CDX, and HOGZILLA).

## CONFLICTS OF INTEREST

The authors declare no conflict of interest.

## AUTHORS

**Asmaa BENCHAMA** obtained her Master's degree in Networks and Systems from ENSA Marrakech, Cadi Ayyad University, Marrakech, Morocco. Presently, she is a research scholar at the Faculty of Science in Agadir, Ibnzohr University, Morocco, and is actively pursuing her Ph.D. in Cyber Security. Her main research focuses include Network Security, Cyber Security, and Artificial Intelligence.

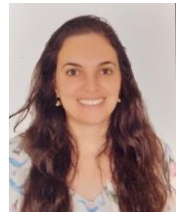

**DR Khalid ZEBBARA** He obtained his Ph.D. in Computer Systems from Ibnzohr University in Agadir, Morocco. Currently, he serves as a Professor at the Faculty of Science AM, Ibnzohr University, Agadir. Additionally, he leads the research team known as Imaging, Embedded Systems, and Telecommunications (IMIS) at the Faculty of Science AM, Ibnzohr University, Agadir.

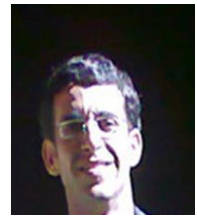